# Thermal Human face recognition based on Haar wavelet transform and series matching technique

Ayan Seal, Suranjan Ganguly, Debotosh Bhattacharjee, Mita Nasipuri, Dipak kr. Basu

Department of Computer Science and Engineering, Jadavpur University, India, Kolkata-700032
ayan.seal@gmail.com, suranjanganguly@gmail.com, debotosh@indiatimes.com, mnasipuri@cse.jdvu.ac.in, dipakkbasu@gmail.com

**Abstract.** Thermal infrared (IR) images represent the heat patterns emitted from hot object and they don't consider the energies reflected from an object. Objects living or non-living emit different amounts of IR energy according to their body temperature and characteristics. Humans are homoeothermic and hence capable of maintaining constant temperature under different surrounding temperature. Face recognition from thermal (IR) images should focus on changes of temperature on facial blood vessels. These temperature changes can be regarded as texture features of images and wavelet transform is a very good tool to analyze multi-scale and multi-directional texture. Wavelet transform is also used for image dimensionality reduction, by removing redundancies and preserving original features of the image. The sizes of the facial images are normally large. So, the wavelet transform is used before image similarity is measured. Therefore this paper describes an efficient approach of human face recognition based on wavelet transform from thermal IR images. The system consists of three steps. At the very first step, human thermal IR face image is preprocessed and the face region is only cropped from the entire image. Secondly, "Haar" wavelet is used to extract low frequency band from the cropped face region. Lastly, the image classification between the training images and the test images is done, which is based on low-frequency components. The proposed approach is tested on a number of human thermal infrared face images created at our own laboratory and "Terravic Facial IR Database". Experimental results indicated that the thermal infra red face images can be recognized by the proposed system effectively. The maximum success of 95% recognition has been achieved.

## 1 Introduction

Since last three decades there exists many commercially available systems of face recognition technology to identify human faces; however face recognition is still a challenging area in computer vision and pattern recognition. The objectives of the face recognition system is to match the data i.e. faces in the stored database to determine the identity of the possible candidate. Face recognition is a sophisticated problem because of the generally similar shape of faces shared with the numerous variations between images of the same face. Various methods have been used to solve the problem. Every method has its own merits and demerits. Most of the research works in this area have been focused on visible spectrum imaging due to easy availability of low cost visible band optical cameras. But, it requires an external source of illumination. Even though the success of automatic face recognition techniques in many practical applications, the task of face recognition based only on the visible spectrum is still a challenging problem under uncontrolled environments. Thermal IR images [1] have been suggested as a possible alternative in handling situations where there is no control over illumination. Thermal IR images represent the heat patterns emitted from an object and they don't consider the reflected energy. Objects emit different amounts of IR energy according to their body temperature and characteristics. Previously, Thermal IR camera was costly but recently the cost of IR cameras has been considerably reduced with the development of CCD technology [2]; thermal images can be captured under different lighting conditions, even under completely dark environment. Using thermal images, the tasks of face detection, localization, and segmentation are comparatively easier and more reliable than those in visible band images [3]. Humans are homoeothermic and hence capable of maintaining constant temperature under different surrounding temperature and since, blood vessels transport warm blood throughout the body; the thermal patterns of faces are derived primarily from the pattern of blood vessels under the skin. These temperature changes can be regarded as texture features of images and wavelet transform is a very good tool to analyze texture with multi scales and multiple directions. In the recent years, wavelet analysis is being popular to the researchers in the field of both theoretical and applied mathematics, and the wavelet transform in particular has established to be an effective tool for data analysis, numerical analysis, image processing [4] etc. Face recognition is realized by image or feature comparison. The pixels of the whole image are concatenated in row major order or column major order. So, the image can be viewed as a series or vectors or sequences. The problem of image comparison can be transformed into the problem of series comparison. The size of the facial images are generally larger, so the number of pixels of facial images are usually huge, so the wavelet transform is used before image comparison, which can effectively reduce the computational complexity. The paper is arranged as follows. Section 2 presents about the outline proposed system. Section 3 shows the experiment and results. Finally, Section 4 concludes and mentions some remarks about different aspects analyzed in this paper.

## 2   Outline of the proposed system

The proposed Thermal Face Recognition System (TFRS) can be subdivided into four main parts, namely image acquisition, image preprocessing, feature extraction, and classification. In image acquisition stage, a FLIR 7 thermal infrared camera has been used to acquire 24-bits colour thermal face images. The images are saved in JPEG format. A thermal face image depicts interesting thermal information of a facial model. The image pre-processing part involves binarization of the acquired thermal IR image, extraction of largest component as the face region, finding the centroid of the face region and finally cropping of the face region in elliptical shape. In feature extraction part, finds LL band, HL band, LH band and HH band using "Haar" wavelet transform. The LL band contains sufficient information to represent the original image and all other band are to be eliminated here. However, size of the LL band is one fourth of the original image. So, "Haar" wavelet is used to reduce dimensionality of the original image. Two dimensional LL band image are converted into one-dimensional horizontal vector in row-major order. Then these reduced one-dimensional horizontal feature vectors are fed into a series matching classifier. The block diagram of the proposed system is given in Fig. 1. Different image processing and classification techniques used here are discussed in detail in subsequent subsections.

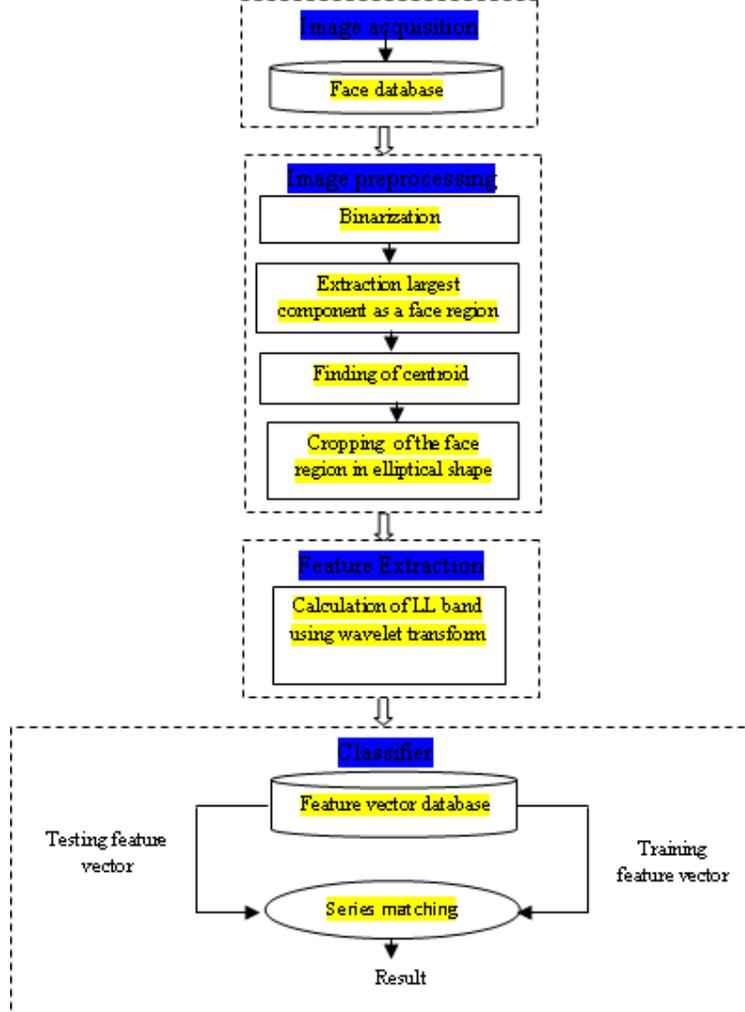

**Fig. 1.** Schematic block diagram of the proposed system

### 2.1 Thermal face image acquisition

In the present work, thermal and visible face images are acquired simultaneously under variable expressions, poses and with/without glasses. Till now 76 individuals have volunteered for this photo shoots and for each individual 39 different templates of RGB color images with Exp1 (happy), Exp2 (angry), Exp3 (sad), Exp4 (disgusted), Exp5 (neutral), Exp6 (fearful) and Exp7 (surprised) are taken. Different pose changes about x-axis, y-axis and z-axis are also taken. Resolution of each image is 320×240 and the images are saved in JPEG format. Two different cameras are used to capture this database. One is Thermal – FLIR 7 and another is Visible – Sony cyber shot. A typical thermal face image is shown in Fig. 2a). This thermal face image depicts interesting thermal information of a facial model.

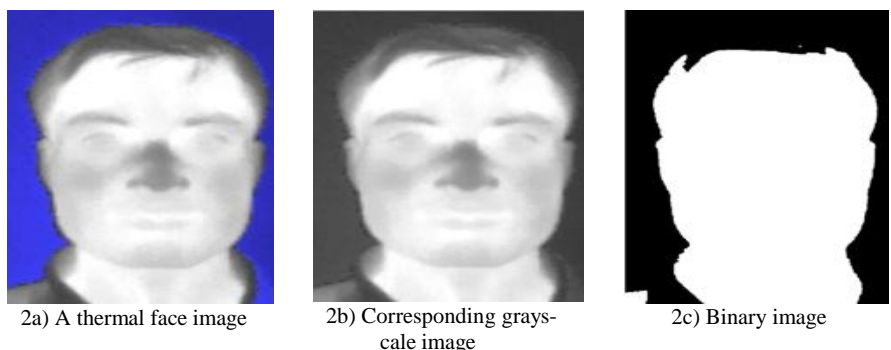

2a) A thermal face image    2b) Corresponding grayscale image    2c) Binary image

**Fig 2.** Thermal face image and its various preprocessing stages

### 2.2 Binarization

Each of the captured 24-bits colour images have been converted into its 8-bit grayscale image. The grayscale image from the previous sample image is shown in Fig. 2b). Then convert those grayscale images into corresponding binary images. The resultant image replaces all pixels in the grayscale image with luminance greater than mean intensity with the value 1 (white) and replaces all other pixels with the value 0 (black). In binary image, black pixels mean background and white pixels mean the face region. The corresponding binary image of the image given in Fig. 2b) is shown in Fig. 2c).

### 2.3 Extracted largest component

The foreground of a binary image may contain more than one object. Let us consider image, in Fig. 2c), it has three objects or components. The large one represents the face region. The others are at the left bottom corner and small dot on the top. Then largest component has been extracted from binary image using "Connected Component Labeling" algorithm [5]. This algorithm is based either on "4-conneted" neighbours or "8-connected" neighbours method [6]. As an illustrative example, consider a largest component as a face skin region illustrated in Fig. 3 using "Connected component labeling" algorithm. It is a binary image. Here, white means face skin region representing with "1" and black means background representing with "0".

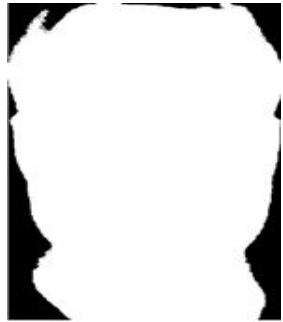 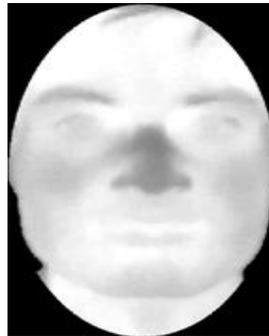

**Fig. 3.** A largest component as a face skin region      **Fig. 4.** Face region in elliptic shape

### 2.4 Finding the centroid [7]

Centroid has been extracted from the binary image using equation 1 and 2.

$$X = \frac{\sum m_{f(x,y)} x}{\sum m_{f(x,y)}} \ldots\ldots\ldots (1) \qquad Y = \frac{\sum m_{f(x,y)} y}{\sum m_{f(x,y)}} \ldots\ldots\ldots (2)$$

Where x, y is the co-ordinate of the binary image and m is the intensity value that is mf(x, y) = f(x, y) =0 or 1.

### 2.5 Cropping of the face region in elliptic shape

Normally human face is of an elliptical shape. Then from the centroid, human face has been cropped in elliptical shape using "Bresenham ellipse drawing" [8] algorithm where, X and Y is x co-ordinate and y-co-ordinate respectively for the centroid which is calculated by equation 1 and 2. Distance between the centroid and the right ear is called the minor axis of the ellipse and distance between the centroid and the forehead is called major axis of the ellipse. After finding the centroid of the face, face has been cropped in elliptical shape and mapped to the grayscale image, which is shown in Fig. 4.

### 2.6 Dimensionality reduction using wavelet transforms

The first discrete wavelet transform (DWT) was invented by the Hungarian mathematician Alfréd Haar in 1909. A key advantage of wavelet transform over Fourier transforms is temporal resolution. Wavelet transform captures both frequency and time i.e. location information. The DWT has a huge number of applications in science, engineering, computer science and mathematics. The Haar transformation is used here for dimensionality reduction since it is the simplest wavelet transform of all and can successfully serve our purpose. Wavelet transform has merits of multi-resolution, multi-scale decomposition, and so on. To obtain the standard decomposition [9] of a 2D image, the 1D wavelet transform to each row is applied first. This operation gives an average pixel value along with detail coefficients for each row. These transformed rows are treated as if they were themselves in an image. Now, 1D wavelet transform to each column is applied. The resulting pixel values are all detail coefficients except for a single overall average coefficient. As a result the elliptical shape facial image is decomposed, and then four regions can be gained. These regions are one low-frequency $LL_1$ (approximate component), and three high-frequency region, namely $LH_1$ (horizontal component), $HL_1$ (vertical component), and $HH_1$ (diagonal component), respectively. The low frequency sub-band $LL_1$ can be further decomposed into four sub-bands $LL_2$, $LH_2$, $HL_2$ and $HH_2$ at the next coarse scale. $LL_i$ is a reduced resolution corresponding to the low-frequency part of an image. The sketch map of the quadratic wavelet decomposition is shown in Fig. 5.

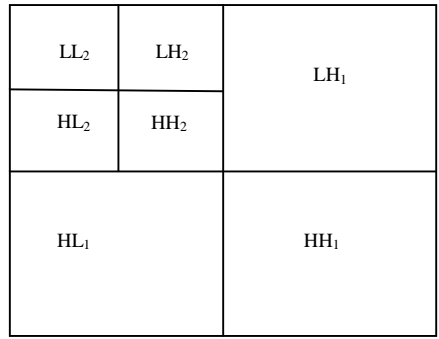

**Fig. 5.** Sketch map of the quadratic wavelet decomposition

As illustrated in fig. 5, the L denotes low frequency and the H denotes high frequency, and that subscripts named from 1 to 2 denote simple, quadratic wavelet decompositions respectively. The standard decomposition algorithm is given below:-

```
function StandardDecomposition(Im[1 to r,1 to c])

// Im[1 to r,1 to c] is an image realized by 2D array,
where r is the number of rows and c is the number of
column.
          for i=1 to r

            1D wavelet transforms (row-number (i))

          end

          for j=1 to c

            1D wavelet transforms (column-number (j))

          end

   end
```

Let's start with a simple example of 1D wavelet transform [10]. Suppose an image with only one row of four pixels, having intensity values [10 4 9 5]. Now apply the Haar wavelet transform on this image. To do so, first pair up the input intensity values or pixel values, storing the mean in order to get the new lower resolution image with intensity values [7 7]. Obviously, some information might be lost in this averaging process. Some detail coefficients need to store to recover the original four intensity values from the two mean values, which capture the missing information. In this example, 3 is the first detail coefficient, since the computed mean is 3 less than 10 and 3 more than 4. This single number is responsible to recover the first two pixels of original four-pixel image. Similarly, the second detail coefficient is 2. Thus, the original image is decomposed into a lower resolution (two-pixel) version and a pair of detail coefficients. Repeating this process recursively on the averages gives the full decomposition, which is shown in Table 1:

**Table 1.** Resolution, mean and the detail coefficients of full decomposition.

| Resolution | Mean | Detail coefficients |
|---|---|---|
| 4 | [10 4 9 5] | |
| 2 | [7 7] | [3 2] |
| 1 | [7] | [0] |

Thus, the one-dimensional Haar wavelet transform of the original four-pixel image is given by [7 0 3 2]. First, 1D wavelet transforms (StandardDecomposed (image) algo-rithm) is used on Fig. 4 row-wise manner, the resultant figure is shown in Fig. 6a). Secondly, 1D wavelet transforms (Standard Decomposed (image) algorithm) is used on Fig. 6a) column-wise manner and the resultant image is shown in Fig. 6b)

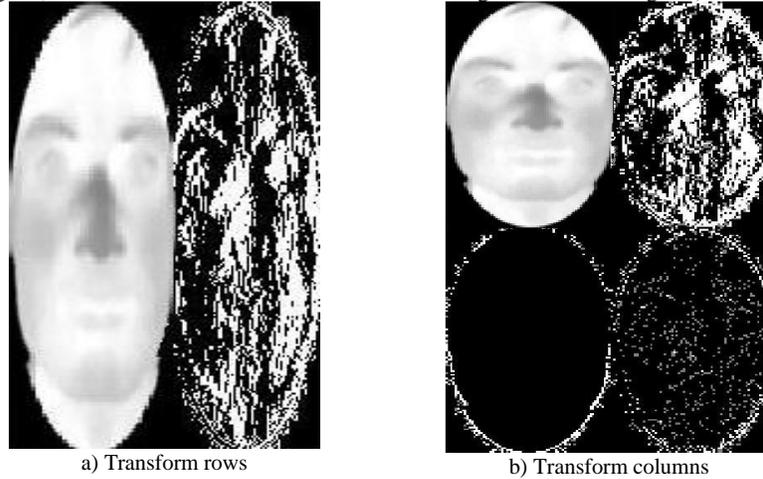

a) Transform rows  b) Transform columns

**Fig. 6.** Haar wevlet transform

The pixels of $LL_2$ image can be concatenated row wise (horizontally) or column wise vertically manner. So the image can be treated as a series. Now the problem of image comparison converts into series comparison.

### 2.7 Series matching classifier [11]

In a gray scale image, each and every pixel is represented by 8 bits, and the range of intensity value is between 0 to 255. Suppose that $l_1 = (l_{11}, l_{12}, l_{13}, ..., l_{1k})$ and $l_2 = (l_{21}, l_{22}, l_{23}, ..., l_{2k})$ are two numerical series, where $0 \leq l_{ij} \leq 255$, $1 \leq i \leq 2$ and $1 \leq j \leq k$. Then the similarity between $l_1$ and $l_2$ can be written as

$$sim(l_1, l_2) = \sum_{j=1}^{k} f(l_1(j), l_2(j)) \quad \dots \dots \dots \dots (3)$$

where $f(.)$ is the matching or comparison function used for similarity measurement. $l_1(j)$ and $l_2(j)$ are the jth elements of each series. The comparison function or matching function is defined as

$$f(l_1(j), l_2(j)) = |l_1(j) - l_2(j)| \quad \dots \dots \dots \dots (4)$$

The matching or comparison having maximal similarity or match is regarded as the optimal comparison, and the corresponding comparison similarity is called as similarity between two series. Images will be classified based on the equation 3 and 4, which is known as series matching classifier.

## 3   Experiment and results

Experiments have been performed on UGC-JU thermal face database, created at our own laboratory and Terravic Facial IR database [12] as well. Till now 76 individuals have volunteered for this photo shoots and for each individual 39 different templates of RGB color images with different pose and different facial expression changes like happy, surprise, fear, etc. Thermal IR image doesn't depend on surrounding lighting conditions since thermal infrared camera capture only emitted energy i.e. temperature from an object. The details of our database have been discussed in section 2.1. The Terravic facial IR database is composed of 20 persons, each of which has different number of images of 320×240 pixels. 10 images of each individual have been selected first. So, 200 images are used for our experiment. If these original images are directly used, the range of value of $sim(l_1,l_2)$ is too large. For example, according to the equation (3) and (4), when two images are completely different, then their $sim(l_1,l_2)$ value is 320×240×255=19584000. Whereas, when two images are completely same, their $sim(l_1,l_2)$ value is regarded as 0. If the image size is large, more computation is required to determine whether two images are same. Therefore, wavelet transform is used to decompose and reduce original images before image comparison. Different orthogonal wavelet filters are available such as Haar, Daubechies, Coiflets, Symlets, etc. Haar wavelet is used here due to simplicity. It is simple and popular as compared to the other wavelet filters. The sizes of the images are reduced significantly after second label wavelet transform. Firstly, the decomposition using Haar wavelet transform is done on the original image and secondly on the low frequency component. LL coefficients are only used here because detail coefficients (LH, HL, HH) may be more useful as features in face recognition to increase the success rate but that will increase processing time enormously. After second label wavelet transform, each image is transformed into horizontal vector, by concatenating pixel in row-wise manner, to form an M×N 2D array (M is the number of images and N is the number of feature vector of images each), where each row corresponds to an image. Clearly, each series has the same length. Each series is composed of only intensity values between 0 to 255 including 0 and 255. Then equation (3) and (4) can be used to compare two images and the corresponding $sim(l_1,l_2)$. Before image comparison, the whole M×N array is divided into two parts of size (M/2) ×N. The odd numbers of rows are taken from the original matrix and put them into the first matrix. Then even numbers of rows are taken from the original matrix and put them into the second matrix. This matrix is used for testing purpose.  Then average of each rows of the training set is calculated column-wise to form a 1D array or series. Let's say this series or array is named as 'X'. Then this 1D array or series is used to differentiate them with each row of the training set using equation (4) and sum of all difference of each row is stored in another 1D array, named as 'Y' using equation (3). That means $sim(l_1,l_2)$ is measured using equation (3) and (4). The above process is also used for testing sets and this series or 1D array is named as 'Z'. Finally, pick one element from 'Z' and find the closest element among 'Y' and then based on this closest element of 'Y', it would be classified. The obtained experimental result is shown in Table 2.

Table 2. Performance rate for different databases

| Name of the database | Label | Recognition rate (%) |
|---|---|---|
| UGC-JU thermal IR database | Original image | 85 |
|  | LL1 | 91 |
|  | LL2 | 95 |
| Terravic Facial IR Database | Original image | 84 |
|  | LL1 | 92 |
|  | LL2 | 93 |

Table 2 presents original image and two different decomposing labels, decomposed by Haar wavelet transform and their recognition rates. Results of face recognition, which are obtained by LL2 component, are better than other components for UGC-JU thermal face database and Terravic Facial IR database, using comparison of each series. The number of pixels of LL2 component is fewer than LL1 component, so that matching time of LL2 components are significantly reduced and the higher recognition rate is also obtained at the same time. A few experimental results based on thermal face images are being given in Table 3, for understanding that our approach is simple and good enough to recognize a person easily.

Table 3. A comparative study based on performance of different thermal face recognition methods (adapted from [13])

| Method | Recognition rate (%) |
|---|---|
| Segmented Infrared Images via Bessel forms [14] 2004) | 90 |
| PCA for Visual indoor Probes [15] | 81.54 |
| PCA+LWIR( Indoor probs) [15] | 58.89 (Maximum) |
| LDA+LWIR (Indoor probs) [15] | 73.92 (Maximum) |
| Equinox +LWIR (Indoor probs) [15] | 93.93 (Maximum) |
| PCA+LWIR( Outdoor probs) [15] | 44.29 (Maximum) |
| LDA+LWIR (Outdoor probs) [15] | 65.30 (Maximum) |
| Equinox+LWIR (Outdoor probs) [15] | 83.02 (Maximum) |
| Eigenfaces +LWIR (Different illumination but same expression) [16] | 95.0 (Average), 89.4 (Minimum) |
| Eigenfaces +LWIR(Different illumination and expression) [16] | 93.3 (Average), 86.8 (Minimum) |

## 4 Conclusion

The proposed human thermal face recognition based on Haar wavelet transform and series matching has been introduced and implemented. The proposed system gave higher recognition rate in the experiments. One of the major advantages of this approach is the ease of implementation. Furthermore, no knowledge of geometry or specific feature of the face is required. However, this system is applicable to front views and constant background only. It may fail in unconstraint environments like natural scenes.

## Acknowledgements

Authors are thankful to a major project entitled "Design and Development of Facial Thermogram Technology for Biometric Security System," funded by University Grants Commission (UGC),India and "DST-PURSE Programme" at Department of Computer Science and Engineering, Jadavpur University, India for providing necessary infrastructure to conduct experiments relating to this work. Ayan Seal is grateful to Department of Science & Technology (DST), India for providing him Junior Research Fellowship-Professional (JRF-Professional) under DST-INSPIRE Fellowship programme [No: IF110591].